\pdfoutput=1

\documentclass[11pt]{article}

\usepackage[final]{acl}

\usepackage{times}
\usepackage{latexsym}

\usepackage[T1]{fontenc}

\usepackage[utf8]{inputenc}

\usepackage{microtype}

\usepackage{inconsolata}

\usepackage{graphicx}
\usepackage{placeins}
\usepackage{float}
\usepackage{amsmath}
\usepackage{booktabs}
\usepackage{CJKutf8}
\usepackage{threeparttable}
\usepackage{multirow}

\title{Rethinking Sign Language Translation: The Impact of Signer Dependence on Model Evaluation}

\author{
  \textbf{Keren Artiaga\textsuperscript{1}},
  \textbf{Sabyasachi Kamila\textsuperscript{2}},
  \textbf{Haithem Afli\textsuperscript{1}},\\
  \textbf{Conor Lynch\textsuperscript{3}},
  \textbf{Mohammed Hasanuzzaman\textsuperscript{1,4}}\\
  \\
  \textsuperscript{1}ADAPT Centre, Munster Technological University, Cork, Ireland\\
  \textsuperscript{2}Manipal Institute of Technology Bengaluru, MAHE, Manipal, India\\
  \textsuperscript{3}Nimbus Research Centre, Munster Technological University, Cork, Ireland\\
  \textsuperscript{4}EEECS, Queen’s University Belfast, UK\\
  \small{
    \textbf{Correspondence:} \href{mailto:keren.artiaga@adaptcentre.ie}{keren.artiaga@adaptcentre.ie}
  }
}

\begin{document}
\maketitle

\begin{abstract}

Sign Language Translation has advanced with deep learning, yet evaluations remain largely signer-dependent, with overlapping signers across train/dev/test. This raises concerns about whether models truly generalise or rely on signer-specific regularities. We perform signer-fold cross-validation on GFSLT-VLP, GASLT, and SignCL, three leading, publicly available, gloss-free SLT models, on CSL-Daily and PHOENIX14T. Under signer-independent evaluation, performance drops sharply: in PHOENIX14T, GFSLT-VLP falls from BLEU-4 21.44 to 3.59 and ROUGE-L 42.49 to 11.89; GASLT from 15.74 to 8.26; and SignCL from 22.74 to 3.66. We also observe that in CSL-Daily, multiple signers perform many target sentences, so common splits can place identical sentences in both training and test, inflating absolute scores by rewarding recall of recurring sentences rather than genuine generalisation. These findings indicate that signer-dependent evaluation can substantially overestimate SLT capability. We recommend: (1) adopting signer-independent protocols to ensure generalisation to unseen signers; (2) restructuring datasets to include explicit signer-independent, sentence-disjoint splits for consistent benchmarking; and (3) reporting both signer-dependent and signer-independent results together with train–test sentence overlap to improve transparency and comparability.

\end{abstract}

\begingroup
\renewcommand\thefootnote{}
\footnotetext{Published in \textit{Findings of the Association for Computational Linguistics: EMNLP 2025}. DOI: 10.18653/v1/2025.findings-emnlp.997}
\addtocounter{footnote}{-1}
\endgroup

\section{Introduction}

Sign Language Translation (SLT) aims to convert sign language videos into spoken or written language, enabling communication between deaf and hearing communities. A key challenge in SLT is signer dependence, where models are often trained and evaluated on splits where the same signers appear in both training and test sets. This setup risks overestimating model performance, as models may exploit signer-specific patterns rather than learning to generalise to unseen individuals \cite{LIU2024127479, 10.1371/journal.pone.0273649, Baytas2024SignerIndependent}.

The most widely used SLT benchmark, RWTH-PHOENIX-Weather-2014T (Phoenix14T) \cite{phoenix14T} \footnote{RWTH-PHOENIX-Weather-2014T is distributed by the HLT \& Pattern Recognition Group, RWTH Aachen University, Germany. The dataset is publicly available for research use}, exemplifies this problem. Its default split includes overlapping signers across train, development, and test sets. Although it provides high-quality gloss and translation annotations across ~8,000 weather forecast videos featuring nine signers, evaluations on this split do not reflect signer-independent performance. A similar issue exists in CSL-Daily \cite{zhou2021improving} \footnote{CSL-Daily is released by the Visual Sign Language Research Group, USTC, under a non-exclusive, non-transferable agreement. Access requires signing a license restricting use to non-commercial academic research, prohibiting redistribution or commercial use, and preserving subject anonymity.}, a large-scale Chinese SLT dataset covering daily topics with 10 signers, which has also become a standard benchmark \cite{Zhou_2023_ICCV, chen2024factorized, wong2024signgpt}.

Many recent SLT models—including both gloss-based \cite{yao2023sign} and gloss-free approaches \cite{Zhou_2023_ICCV, ye2024improving, chen2024factorized, wong2024signgpt, gong2024llms}—have reported impressive gains. However, these gains are typically measured on signer-overlapping splits, leaving open the question of how well such models generalise to unseen signers.

To address this, we perform signer-fold cross-validation on Phoenix14T and CSL-Daily, evaluating three state-of-the-art gloss-free SLT models: GFSLT-VLP \cite{Zhou_2023_ICCV}, GASLT \cite{yin2023gloss}, and SignCL \cite{ye2024improving}. Each fold withholds a signer entirely from training and development. Our results (Section~\ref{sec:results}) show that, on average, translation performance drops under signer-independent evaluation compared to default-split baselines. This highlights how signer-dependent evaluations can obscure the true generalisation ability of SLT models.

This study provides the first systematic analysis of signer-independent SLT performance using signer-fold cross-validation. We advocate for future SLT research to adopt signer-independent protocols to ensure robust and fair evaluation. The remainder of this paper reviews related work, describes the experimental setup, presents results and qualitative analyses, and discusses implications for future SLT development.

\section{Related Works}

Signer independence remains a critical challenge in Sign Language Recognition (SLR) and Sign Language Translation (SLT), where models often overfit to signer-specific traits such as hand shape, articulation, or signing speed. While this issue has been well-documented in SLR since early work in 2013 \cite{10.1007/978-3-642-53932-9_26}, it has received comparatively little attention in SLT.

Most existing SLT research—whether gloss-based or gloss-free—continues to evaluate models using signer-overlapping splits. Gloss-based pipelines, including Gloss-to-Text (G2T), Sign-to-Gloss → Gloss-to-Text (S2G→G2T), and related variants \cite{camgoz2018neural, yin-read-2020-better, yin2021simulslt, kan2022sign}, have shown steady performance improvements, but do not assess generalisation to unseen signers. This limitation persists in recent models such as IP-SLT \cite{yao2023sign} and CV-SLT \cite{Zhao_2024_AAAI}, which continue to rely on signer-dependent splits.

Gloss-free SLT approaches, which bypass intermediate gloss representations, have gained traction for their scalability in low-resource settings. Early models such as S2T \cite{camgoz2020sign}, NSLT \cite{Orbay2020NST}, and TSPNet \cite{NEURIPS2020_8c00dee2} laid the groundwork, but similarly evaluated only on signer-overlapping data. Recent vision-language pretrained models—GFSLT-VLP \cite{Zhou_2023_ICCV}, GASLT \cite{yin2023gloss}, SignCL \cite{ye2024improving}, Sign2GPT \cite{wong2024signgpt}, and SignLLM \cite{gong2024llms}—have reported strong results on Phoenix14T and CSL-Daily. However, these evaluations remain constrained to the default splits, which do not isolate signer effects.

Although signer-dependent training has been shown to inflate performance in SLR \cite{s23167156}, SLT studies have yet to investigate this systematically. No prior work, to our knowledge, has performed signer-fold cross-validation in SLT to evaluate robustness across all signers. Our study addresses this gap by applying signer-fold evaluation to three strong, publicly available gloss-free models: GFSLT-VLP, GASLT, and SignCL. This allows us to quantify how signer overlap in standard protocols may obscure the true generalisation ability of SLT systems.

\section{Experiments and Results}

\subsection{Methodology}

The default distribution of Phoenix14T consists of 7,022 videos in the training set, 269 videos in the development set and 966 videos in the test set - with nine signers overlapping across these splits. While this setup facilitates training, it allows models to exploit signer-specific features such as hand shape, and signing style, rather than learning generalisable representations for unseen signers.

To address this issue, signer-fold cross-validation was applied to the Phoenix14T dataset. Unlike the default split, signer-fold cross-validation ensures that no signers are shared across training, development, or test sets. The dataset was divided into nine folds, with each fold containing videos from one signer exclusively used for testing, another for development, and the remaining signers for training. The size of the training set varied across folds, ranging from 5,100 to 7,893 videos, as shown in Table~\ref{tab:signer_fold_results}. This setup provides a robust framework for evaluating how well models generalise to unseen signers. We report both fold-specific results for each test signer and aggregate results computed by averaging scores across all folds.

Similarly, the CSL-Daily dataset was reorganised to support signer-independent evaluation. Ten signer folds were created, with training set sizes ranging from 12,490 to 18,338 videos, development sets from 771 to 7,391 videos, and test sets from 771 to 1,978 videos, as shown in Table~\ref{tab:csl_fold_results}. For this dataset as well, we report both per-fold scores and mean aggregate scores across all ten folds.

For SignCL and GFSLT-VLP, training each signer-based fold on PHOENIX14T took approximately 6–7 days, while on CSL-Daily it took around 14 days, depending on the dataset size for each fold. The models were trained on an NVIDIA Quadro RTX 6000 GPU with 24 GB of dedicated VRAM. To ensure consistency and comparability with prior work, all hyperparameter values were retained from the original implementations of the models. For GASLT, training across all folds on PHOENIX14T took 2–3 days using the same type of GPU (NVIDIA Quadro RTX 6000) with 24 GB of dedicated VRAM.

To assess model performance, corpus-level BLEU-4 and ROUGE-L were employed. All models were evaluated using their original implementations: SignCL and GFSLT-VLP report corpus-level BLEU and ROUGE scores using the \texttt{nlg-eval}~\cite{sharma2017nlgeval} package, while GASLT uses the \texttt{sacreBLEU}~\cite{post-2018-call} package for corpus-level BLEU and the \texttt{mscoco\_rouge}\footnote{\url{https://github.com/tylin/coco-caption/tree/master/pycocoevalcap}} package for ROUGE-L.
.

\paragraph{BLEU-4} (Bilingual Evaluation Understudy) measures the precision of n-grams between the generated and reference translations while applying a brevity penalty to prevent overly short outputs \cite{papineni2002bleu}. The BLEU score is computed as:

\begin{equation}
    \text{BLEU} = BP \cdot \exp \left( \sum_{n=1}^{4} w_n \log p_n \right)
\end{equation}

where \( p_n \) represents the precision of n-grams up to length 4, \( w_n \) is the weight assigned to each n-gram, and \( BP \) is the brevity penalty defined as:

\begin{equation}
    BP = 
    \left\{
    \begin{array}{ll}
        1, & \text{if } c > r \\
        e^{(1 - r/c)}, & \text{if } c \leq r 
    \end{array}
    \right\}
\end{equation}

where \( c \) is the length of the generated translation and \( r \) is the length of the reference translation.

\paragraph{ROUGE-L} (Recall-Oriented Understudy for Gisting Evaluation) evaluates translation quality based on the longest common subsequence (LCS) between the generated and reference sentences \cite{lin-2004-rouge}. The ROUGE-L score is computed as:

\begin{equation}
    \text{ROUGE-L} = \frac{LCS(X, Y)}{\max(|X|, |Y|)}
\end{equation}

where \( LCS(X, Y) \) represents the longest common subsequence between the candidate translation \( X \) and the reference \( Y \), and \( |X| \) and \( |Y| \) denote their respective lengths.

\subsection{Results and Analysis}
\label{sec:results}

\subsubsection{Results on Phoenix14T}

Table~\ref{tab:signer_fold_results} shows a clear performance gap between the default and signer-independent settings. GFSLT-VLP, for example, drops from 21.44 BLEU-4 and 42.49 ROUGE-L (default) to an average of 10.53 BLEU-4 and 26.14 ROUGE-L across signer folds, a relative drop of over 50\%. GASLT and SignCL exhibit similar trends, with GASLT falling from 15.74 to 10.24 BLEU-4 and SignCL from 22.74 to just 4.18.

To contextualise these results with respect to sentence overlap, Table~\ref{tab:phoenix_leakage_by_fold} reports sentence-level leakage for PHOENIX14T. Dev/test \emph{unique} sentences overlap only marginally with training (typically $\approx$1–3\% per fold), and the fraction of test sentences performed by $\geq$3 signers is very small. Thus, broad repetition is rare and evaluation inflation from repeated targets is minimal.

Despite the overall degradation, per-signer performance reveals interesting patterns. GFSLT-VLP performs best in Fold~6 (BLEU-4: 17.30) but worst in Fold~8 (BLEU-4: 3.59), suggesting that signer variability—such as articulation style or test-set size—strongly impacts results. GASLT also fluctuates across folds but with smaller standard deviations, indicating more consistent signer-agnostic behaviour. In Fold~3, for instance, GASLT slightly outperforms GFSLT-VLP (10.19 vs.\ 10.02 BLEU-4), hinting at different sensitivities to signer traits. SignCL, while strongest in the default setting, shows the weakest generalisation, averaging just 4.18 BLEU-4 across folds.

These results underscore the importance of evaluating models under signer-independent protocols to reveal robustness gaps that may be hidden in standard splits.

\begin{table*}[!ht]
\centering
\scriptsize
\renewcommand{\arraystretch}{1.2}
\caption{Signer-Fold Cross-Validation Results on PHOENIX14T for GFSLT-VLP, GASLT, and SignCL models}
\label{tab:signer_fold_results}
\setlength{\tabcolsep}{3pt}
\begin{tabular}{@{}cccccccccccc@{}}
\toprule
\textbf{Fold} & \textbf{Dev Signer} & \textbf{Test Signer} & \textbf{Train Size} & \textbf{Dev Size} & \textbf{Test Size} 
& \multicolumn{2}{c}{\textbf{GFSLT-VLP}} 
& \multicolumn{2}{c}{\textbf{GASLT}} 
& \multicolumn{2}{c}{\textbf{SignCL}} \\
\cmidrule(lr){7-8} \cmidrule(lr){9-10} \cmidrule(lr){11-12}
& & & & & & \textbf{BLEU-4} & \textbf{ROUGE-L} & \textbf{BLEU-4} & \textbf{ROUGE-L} & \textbf{BLEU-4} & \textbf{ROUGE-L} \\
\midrule
1 & Signer08 & Signer01 & 5,100 & 966  & 2,191 & 6.65  & 21.80 & 7.94  & 24.46 & 2.58 & 9.79 \\
2 & Signer01 & Signer02 & 5,971 & 2,191 & 95    & 8.49  & 20.61 & 8.89  & 22.75 & 4.65 & 13.74 \\
3 & Signer05 & Signer03 & 5,641 & 1,933 & 683   & 10.02 & 26.70 & 10.19 & 27.10 & 4.81 & 14.95 \\
4 & Signer03 & Signer04 & 6,367 & 683   & 1,207 & 13.70 & 32.30 & 11.02 & 29.21 & 4.22 & 13.22 \\
5 & Signer07 & Signer05 & 5,458 & 866   & 1,933 & 11.90 & 29.70 & 9.79  & 27.45 & 3.66 & 12.76 \\
6 & Signer04 & Signer06 & 7,003 & 1,207 & 47    & 17.30 & 34.02 & 12.65 & 31.33 & 5.43 & 15.50 \\
7 & Signer06 & Signer07 & 7,344 & 47    & 866   & 9.19  & 26.30 & 8.26  & 25.40 & 3.05 & 12.13 \\
8 & Signer09 & Signer08 & 7,022 & 269   & 966   & 3.59  & 11.80 & 10.07 & 27.55 & 4.40 & 13.19 \\
9 & Signer02 & Signer09 & 7,893 & 95    & 269   & 13.95 & 32.07 & 13.38 & 30.87 & 4.79 & 13.16 \\
\midrule
\multicolumn{6}{r}{\textbf{Mean ± Std (9 folds)}} 
& \textbf{10.53 ± 4.17} & \textbf{26.14 ± 7.10} 
& \textbf{10.24 ± 1.86} & \textbf{27.35 ± 2.85} 
& \textbf{4.18 ± 0.92} & \textbf{13.16 ± 1.64} \\

\midrule
\multicolumn{3}{l}{\textbf{Default Split}} 
& \textbf{7,096} & \textbf{519} & \textbf{642} 
& \textbf{21.44} & \textbf{42.49} & \textbf{15.74} & \textbf{39.86} & \textbf{22.74} & \textbf{49.04} \\
\bottomrule
\end{tabular}

\vspace{0.5em}
\raggedright
\end{table*}



\begin{table*}[!ht]
\centering
\scriptsize
\renewcommand{\arraystretch}{1.15}
\caption{Sentence leakage per signer-fold on PHOENIX14T.}
\label{tab:phoenix_leakage_by_fold}
\setlength{\tabcolsep}{6pt}
\resizebox{\textwidth}{!}{%
\begin{tabular}{@{}ccccccccc@{}}
\toprule
\textbf{Fold} & \textbf{Dev} & \textbf{Test} & \textbf{Dev overlap} & \textbf{Test overlap} &
\textbf{Dev leak (sent)} & \textbf{Test leak (sent)} & \textbf{Dev+Test leak (sent)} & \textbf{$\ge$3 signers} \\
\midrule
1 & Signer08 & Signer01 & 15 & 31 & 1.59\% & 1.47\% & 1.12\% & 0.86\% \\
2 & Signer01 & Signer02 & 32 &  3 & 1.52\% & 3.26\% & 1.46\% & 3.26\% \\
3 & Signer05 & Signer03 & 21 & 15 & 1.11\% & 2.23\% & 1.18\% & 1.49\% \\
4 & Signer03 & Signer04 & 12 & 17 & 1.79\% & 1.42\% & 1.34\% & 1.00\% \\
5 & Signer07 & Signer05 & 18 & 19 & 2.15\% & 1.01\% & 1.00\% & 0.74\% \\
6 & Signer04 & Signer06 & 20 &  1 & 1.67\% & 2.22\% & 1.69\% & 0.00\% \\
7 & Signer06 & Signer07 &  1 & 20 & 2.22\% & 2.39\% & 2.38\% & 1.56\% \\
8 & Signer09 & Signer08 &  7 & 16 & 2.60\% & 1.70\% & 1.65\% & 1.48\% \\
9 & Signer02 & Signer09 &  3 &  7 & 3.26\% & 2.60\% & 2.77\% & 1.86\% \\
\bottomrule
\end{tabular}%
}

{\raggedright \footnotesize Rates are percentages of \emph{unique} dev/test sentences that also appear in train. “Dev+Test leak (sent)” is computed over the union of dev and test unique sentences. “$\ge$3 signers” is the fraction of test sentences performed by at least three signers \par }
\end{table*}

\subsubsection{Qualitative Analysis for Phoenix14T}

\label{sec:Qualitative_Analysis}

\paragraph{Analysis on the representative fold}
To further illustrate model behavior under signer-independent conditions, we present a detailed set of qualitative comparisons in Table~\ref{tab:fold3_comparison}. This table focuses on multiple examples from Fold 3 (Signer03), a signer selected for its representative performance across models. Fold 3 is near the mean performance for GFSLT-VLP and GASLT, and ranks fourth for SignCL, with a squared error of 0.616—only slightly higher than its best-performing folds (Fold 4: 0.0055, Fold 8: 0.0508, Fold 9: 0.3762). As such, it offers a balanced setting for controlled evaluation of model outputs.

\begin{table*}[!ht]
\centering
\scriptsize
\renewcommand{\arraystretch}{1.3}
\caption{Qualitative comparison of GFSLT-VLP, GASLT, and SignCL predictions on weather forecast references from Fold 3 (Signer03) for Phoenix14T}
\label{tab:fold3_comparison}
\setlength{\tabcolsep}{6pt}
\begin{tabular}{@{}p{4.65cm}p{11cm}@{}}
\toprule
\textbf{Reference Translation} & \textbf{Predictions} \\
\midrule

\textbf{liebe zuschauer guten abend}  
\newline \textit{(Dear viewers, good evening)} &
\textbf{GFSLT-VLP:} \textbf{liebe zuschauer guten abend}  
\newline \textit{(Dear viewers, good evening)}  
\newline \textbf{GASLT:} \textbf{liebe zuschauer guten abend}  
\newline \textit{(Dear viewers, good evening)}  
\newline \textbf{SignCL:} \textbf{ihnen einen schönen abend und machen sie es gut}  
\newline \textit{(Have a nice evening and take care)} \\
\midrule

\textbf{morgen kommt dann kräftig bewegung ins wetter mit einem tief das uns kräftige schneefälle bringen wird und stürmischem wind}  
\newline \textit{(Tomorrow, a low-pressure system will bring heavy snowfall and stormy winds)} &
\textbf{GFSLT-VLP:} \textbf{morgen wechseln sich freundliche momente mit dichten wolken ab dabei fallen schauer am nachmittag sind auch einzelne gewitter dabei}  
\newline \textit{(Tomorrow, pleasant moments will alternate with dense clouds, with showers. There will also be isolated thunderstorms in the afternoon.)}  
\newline \textbf{GASLT:} \textbf{morgen bleibt es bei uns noch trocken und auch in den nächsten tagen da wird es ein bisschen kälter und der mitte und im süden zum teil frostig}  
\newline \textit{(Tomorrow it will remain dry and in the next few days it will be a bit colder and partly frosty in the middle and in the south)}  
\newline \textbf{SignCL:} \textbf{am mittwoch ist es windig später vor allem im süden neue schneefälle auch am donnerstag schneit es gebietsweise noch etwas minus elf bis minus ein grad}  
\newline \textit{(On Wednesday it will be windy, later especially in the south new snowfall. On Thursday there will still be some local snowfalls, from –11 to –1 degrees)} \\
\midrule

\textbf{es klart verbreitet auf minus fünfundzwanzig grad können im osten erreicht werden im westen zum teil auch bis minus einundzwanzig grad}  
\newline \textit{(It will clear up widely, with lows of –25°C in the east and –21°C in the west)} &
\textbf{GFSLT-VLP:} \textbf{am tag abgesehen von einigen schleierwolken viel sonne im süden ist es fast wolkenlos nur über die küsten ziehen auch mal dichtere wolken}  
\newline \textit{(During the day, apart from a few wispy clouds, there will be plenty of sun in the south. It will be almost cloudless, with only denser clouds occasionally moving over the coast.)}  
\newline \textbf{GASLT:} \textbf{im westen sind es heute nacht minus fünf grad sonst werte um den gefrierpunkt oder minus vier bis minus}  
\newline \textit{(In the west it is minus five degrees tonight, otherwise temperatures around freezing or minus four to minus)}  
\newline \textbf{SignCL:} \textbf{im norden und nordwesten fällt morgen vormittag gebietsweise noch regen in der mitte regnet es nur hier und da etwas im süden und osten gibt es neben ein paar schauern auch wolkenlücken}  
\newline \textit{(Rain still falls tomorrow morning in the north and northwest; only scattered rain in the center; in the south and east, some gaps in the clouds between showers)} \\
\midrule

\textbf{in der nacht ist es verbreitet klar oder nur gering bewölkt}  
\newline \textit{(Overnight it will be mostly clear or only lightly cloudy)} &
\textbf{GFSLT-VLP:} \textbf{in der nacht ist es meist klar oder nur leicht bewölkt besonders in den flussniederungen in der mitte und des südens wabern wieder ein paar nebelfelder}  
\newline \textit{(At night it will be mostly clear or only slightly cloudy, especially in the river valleys in the middle and south a few patches of fog will appear again)}  
\newline \textbf{GASLT:} \textbf{in der nacht klart es verbreitet auf stellenweise bildet sich nebel oder hochnebelartige bewölkt}  
\newline \textit{(At night it will clear up widely, in places fog or high fog-like clouds will form)}  
\newline \textbf{SignCL:} \textbf{in der nacht regnet es stellenweise etwas zwischen rheinland pfalz und brandenburg}  
\newline \textit{(Overnight it will rain in places between Rhineland-Palatinate and Brandenburg)} \\
\bottomrule
\end{tabular}
\end{table*}

The examples in Table~\ref{tab:fold3_comparison} show clear differences in how closely each model reproduces the reference translations. Across examples, SignCL is the least faithful to the input, often producing content that is either unrelated or directly contradictory to the reference.

For instance, when the reference is a simple greeting—“Dear viewers, good evening”—both GFSLT-VLP and GASLT reproduce this exactly, while SignCL inserts a closing phrase about having a nice evening. This additional material was not present in the input. In another case, the reference describes an approaching low-pressure system bringing heavy snowfall and stormy winds. Instead of capturing this, GFSLT-VLP replaces the event with thunderstorms, GASLT shifts focus to dry and cold conditions, and SignCL includes generic snowfall and temperature values that were not mentioned in the reference.

There are also cases where SignCL introduces weather events that contradict the reference. For example, when the input states that the night will be mostly clear, SignCL claims it will rain in specific regions. In contrast, GFSLT-VLP and GASLT correctly describe clear skies, with some elaboration. Furthermore, GASLT gives a more moderate description that partially aligns with the reference.

In several predictions, SignCL outputs templated phrases such as “and now the weather forecast for tomorrow” even when the reference contains no such framing. These insertions suggest reliance on memorised genre patterns rather than the input itself. GASLT and GFSLT-VLP are less prone to such hallucinations, but still deviate by generalising the weather type or omitting nouns such as region or adverbs such as intensity.

\paragraph{Analysis on the best performing folds}
To gain further insight into the qualitative behaviour of the models, Table~\ref{tab:gfslt_signcl_comparison} presents a side-by-side comparison of translations produced by GFSLT-VLP and SignCL on selected examples from Fold 06, which corresponds to the best-performing fold for both models. Note that GASLT achieves its best performance on a different fold and is analysed separately.

The examples in the table highlight key differences in how GFSLT-VLP and SignCL handle content preservation, topical focus, and hallucination. GFSLT-VLP typically remains aligned with the input, although it sometimes generalises temporal framing or introduces inferred details such as temperature values. In contrast, SignCL frequently outputs broadcast-style templates—such as “und nun die wettervorhersage…” or opening greetings like “guten abend liebe zuschauer”—which are fluent but semantically unrelated to the input. While both models occasionally hallucinate formulaic broadcast language, GFSLT-VLP tends to preserve more of the original meteorological content, whereas SignCL often substitutes it entirely with generic phrasing. These observations suggest that GFSLT-VLP maintains a stronger grounding in the source input, whereas SignCL appears more susceptible to overfitting on surface-level patterns.

\begin{table}[!ht]
\centering
\scriptsize
\renewcommand{\arraystretch}{1.15}
\caption{Qualitative comparison between GFSLT-VLP and SignCL predictions and
reference translations on selected examples from the
best-performing fold (fold 06) for Phoenix14T}
\label{tab:gfslt_signcl_comparison}
\setlength{\tabcolsep}{3pt}
\begin{tabular}{@{}p{3.7cm}p{3.7cm}@{}}
\toprule
\textbf{Reference Translation} & \textbf{Predictions} \\
\midrule

\textbf{im süden beginnt die nacht klar auch sonst lösen sich die meisten wolken auf}  
\newline \textit{(In the south the night begins clear and most of the clouds will also dissolve)} &
\textbf{GFSLT-VLP:} im süden dürfte es wahrscheinlich bis zum abend trocken bleiben  
\newline \textit{(In the south it is likely to remain dry until the evening)}  
\newline \textbf{SignCL:} und nun die wettervorhersage für morgen sonntag den sechsundzwanzigsten november  
\newline \textit{(and now the weather forecast for tomorrow, Sunday the twenty-sixth of November)} \\
\midrule

\textbf{nur im nordwesten fallen später schauer}  
\newline \textit{(only in the northwest will showers fall later)} &
\textbf{GFSLT-VLP:} ihnen noch einen schönen abend und machen sie es gut  
\newline \textit{(Have a nice evening and take care)}  
\newline \textbf{SignCL:} guten abend liebe zuschauer  
\newline \textit{(good evening dear viewers)} \\
\midrule

\textbf{am tag breiten sich die wolken mit einzelnen schauern von der nordsee über die nordwesthälfte aus}  
\newline \textit{(During the day, clouds with occasional showers spread from the North Sea over the northwestern half)} &
\textbf{GFSLT-VLP:} am tag neunzehn grad ganz im norden und zwölf grad in teilen thüringens und sachsens  
\newline \textit{(during the day nineteen degrees in the far north and twelve degrees in parts of Thuringia and Saxony)}  
\newline \textbf{SignCL:} und nun die wettervorhersage für morgen donnerstag den sechsundzwanzigsten november  
\newline \textit{(and now the weather forecast for tomorrow, Thursday, November 26th)} \\
\midrule

\textbf{zwischen pfalz und sachsen bleibt es am längsten freundlich südöstlich davon regnet es zeitweise}  
\newline \textit{(Between Palatinate and Saxony it stays friendly the longest, southeast of that it rains intermittently)} &
\textbf{GFSLT-VLP:} im südosten bleibt es bis dreizehn grad weitgehend trocken  
\newline \textit{(In the southeast it will remain largely dry up to thirteen degrees)}  
\newline \textbf{SignCL:} und nun die wettervorhersage für morgen donnerstag den sechsundzwanzigsten november  
\newline \textit{(and now the weather forecast for tomorrow, Thursday, November 26th)} \\
\bottomrule
\end{tabular}
\end{table}

On the other hand, Table \ref{tab:gaslt_reference_comparison} presents a qualitative comparison between GASLT predictions and the corresponding human reference translations from the best-performing fold (fold 09). The examples highlight a range of model behaviors, including semantic drift (top rows) and accurate reproduction of fixed expressions such as weather report introductions (bottom rows). These cases illustrate GASLT's tendency to occasionally hallucinate or generalise content (e.g., replacing atmospheric descriptions with generic closings), while also showing its reliability on highly formulaic phrases.

\begin{table}[htbp!]
\centering
\scriptsize
\renewcommand{\arraystretch}{1.2}
\caption{Qualitative comparison between GASLT and reference translations on selected examples from the best-performing fold (fold 09). for Phoenix14T}
\label{tab:gaslt_reference_comparison}
\setlength{\tabcolsep}{4pt}
\begin{tabular}{@{}p{3.7cm}p{3.7cm}@{}}
\toprule
\textbf{Reference Translation} & \textbf{GASLT Prediction} \\
\midrule

\textbf{aber so wird es nicht bleiben} \newline \textit{(but it won't stay that way)} &
\textbf{jetzt wünsche ich ihnen noch einen schönen abend} \newline \textit{(Now I wish you a nice evening)} \\
\midrule

\textbf{an den folgetagen bestimmen tiefdruckgebiete das geschehen} \newline \textit{(In the following days, low pressure areas determine what happens)} &
\textbf{am tag ist es zunächst noch freundlich im süden} \newline \textit{(During the day it is initially still pleasant in the south)} \\
\midrule

\textbf{und nun die wettervorhersage für morgen samstag den dreizehnten februar} \newline \textit{(and now the weather forecast for tomorrow, Saturday the thirteenth of February)} &
\textbf{und nun die wettervorhersage für morgen samstag den dreizehnten februar} \newline \textit{(and now the weather forecast for tomorrow, Saturday the thirteenth of February)} \\
\midrule

\textbf{und nun die wettervorhersage für morgen dienstag den siebten juli} \newline \textit{(and now the weather forecast for tomorrow, Tuesday, July 7th)} &
\textbf{und nun die wettervorhersage für morgen dienstag den siebten juli} \newline \textit{(and now the weather forecast for tomorrow, Tuesday, July 7th)} \\
\bottomrule
\end{tabular}
\end{table}

Overall, GASLT performs reliably on stereotyped and repetitive structures, such as introductory phrases, and demonstrates fluency with common weather-related terminology. However, it often diverges from the reference in subtle but meaningful ways—omitting key meteorological details, hallucinating plausible-sounding but incorrect content, or shifting the topical focus.

\paragraph{Analysis on the worst performing folds}
Meanwhile. Table~\ref{tab:gfslt_failures} illustrates key failure modes of GFSLT-VLP on the weakest-performing fold (Fold 08) of Phoenix14T. The most prominent issue is template overgeneration, with the model frequently defaulting to fixed phrases such as “und nun die wettervorhersage für morgen mittwoch den sechsundzwanzigsten november” regardless of the input, indicating memorisation rather than grounded translation. Additionally, predictions often drift semantically from the reference, replacing specific weather events (e.g., stormy gusts) with unrelated or generic descriptions (e.g., cloud cover). The model also exhibits low output diversity, repeatedly generating fallback phrases such as “am tag mal sonne mal wolken …” across distinct inputs. These patterns suggest a reliance on high-frequency patterns at the expense of semantic accuracy.

\begin{table}[!ht]
\centering
\scriptsize
\renewcommand{\arraystretch}{1.15}
\caption{Qualitative comparison between GFSLT-VLP predictions and reference translations on selected examples from the weakest-performing fold (Fold 08) for Phoenix14T}
\label{tab:gfslt_failures}
\setlength{\tabcolsep}{3pt}
\begin{tabular}{@{}p{3.7cm}p{3.7cm}@{}}
\toprule
\textbf{Reference} & \textbf{GFSLT-VLP Prediction} \\
\midrule

\textbf{am dienstag wird es dann verbreitet freundlich}  
\newline \textit{(On Tuesday it will be generally friendly)} &
\textbf{im norden und westen ist es heute nacht schon stark bewölkt}  
\newline \textit{(In the north and west it is already very cloudy tonight)} \\
\midrule

\textbf{liebe zuschauer guten abend}  
\newline \textit{(Dear viewers, good evening)} &
\textbf{guten abend liebe zuschauer}  
\newline \textit{(Good evening, dear viewers)} \\
\midrule

\textbf{bei gewittern sind stürmische böen möglich}  
\newline \textit{(During thunderstorms, stormy gusts are possible)} &
\textbf{im nordwesten ist es meist stark bewölkt}  
\newline \textit{(In the northwest it is mostly cloudy)} \\
\midrule

\textbf{und nun die wettervorhersage für morgen sonntag den siebenundzwanzigsten märz}  
\newline \textit{(and now the weather forecast for tomorrow, Sunday the twenty-seventh of March)} &
\textbf{und nun die wettervorhersage für morgen mittwoch den sechsundzwanzigsten november}  
\newline \textit{(And now the weather forecast for tomorrow, Wednesday November 26th)} \\
\midrule

\textbf{am wochenende sorgen dann noch mildere luft und viel wind gebietsweise für zweistellige plusgrade}  
\newline \textit{(At the weekend, milder air and strong winds will cause double-digit plus temperatures in some areas)} &
\textbf{und nun die wettervorhersage für morgen mittwoch den sechsundzwanzigsten november}  
\newline \textit{(And now the weather forecast for tomorrow, Wednesday November 26th)} \\
\bottomrule
\end{tabular}
\end{table}

For GASLT and SignCL, we qualitatively analyse reference–prediction pairs from Fold 1, which yielded the poorest performance for both models. As shown in Table~\ref{tab:gaslt_signcl_failures}, SignCL displays a consistent failure mode in which nearly all predictions collapse to variations of a single phrase—\textit{``und nun die wettervorhersage für morgen samstag den neunundzwanzigsten november''}—regardless of the input content. This fallback behavior suggests that the model fails to generalise and instead defaults to a memorised or overused template under signer-independent conditions. By contrast, GASLT exhibits more diverse but still problematic error patterns. A common trend is semantic drift, where the prediction diverges from the reference to describe unrelated weather scenarios. Another frequent issue is temporal mismatch, with outputs shifting events to different times of day, or introducing exaggerated weather details not present in the reference. While GASLT avoids collapsing to a single repeated phrase, its tendency to hallucinate or distort key details highlights difficulties in maintaining temporal and semantic fidelity in low-performing folds.

\begin{table}[!ht]
\centering
\scriptsize
\renewcommand{\arraystretch}{1.15}
\caption{Qualitative comparison between GASLT and SignCL predictions and reference translations on selected examples from Fold 01 of Phoenix14T}
\label{tab:gaslt_signcl_failures}
\setlength{\tabcolsep}{6pt}
\begin{tabular}{@{}p{3.7cm}p{3.7cm}@{}}
\toprule
\textbf{Reference Translation} & \textbf{Predictions} \\
\midrule

\textbf{guten abend liebe zuschauer} \newline
\textit{(good evening dear viewers)} 
&
\textbf{GASLT:} das war es für heute schönen abend noch \newline
\textit{(that's it for today, have a lovely evening)} \newline
\textbf{SignCL:} und nun die wettervorhersage für morgen samstag den neunundzwanzigsten november \newline
\textit{(and now the weather forecast for tomorrow, Saturday the twenty-ninth of November)} \\
\midrule

\textbf{tiefer luftdruck bestimmt unser wetter} \newline
\textit{(low air pressure determines our weather)}
&
\textbf{GASLT:} das tief das von großbritannien nach südskandinavien zieht \newline
\textit{(the low that moves from Great Britain to southern Scandinavia)} \newline
\textbf{SignCL:} und nun die wettervorhersage für morgen samstag den neunundzwanzigsten november \newline
\textit{(and now the weather forecast for tomorrow, Saturday the twenty-ninth of November)} 
\\ \midrule

\textbf{und es ist ein bisschen kühler es ist erträglicher es sind fünfundzwanzig bis dreißig grad noch an nord und ostsee naja da haben luft und wasser dieselbe temperatur zwanzig bis zweiundzwanzig grad} \newline
\textit{(and it's a bit cooler, it's more bearable; it's still 25–30°C at the North and Baltic Sea; there the air and water have the same temperature, 20–22°C)}
&
\textbf{GASLT:} aber es ist sehr windig so in den nächsten tagen bis zum wochenende vor allen dingen im norden und an der nordsee da bleibt es mit minus vier bis minus vier grad im norden \newline
\textit{(but it will be very windy in the next few days until the weekend, especially in the north and at the North Sea, where it will remain at around minus four degrees in the north)} \newline
\textbf{SignCL:} und nun die wettervorhersage für morgen samstag den neunundzwanzigsten november \newline
\textit{(and now the weather forecast for tomorrow, Saturday the twenty-ninth of November)}
\\ \midrule

\textbf{im nordosten zum teil noch regen} \newline
\textit{(in the northeast partly still rain)}
&
\textbf{GASLT: sonst ist es meist stark bewölkt}\newline
\textit{(otherwise it is mostly cloudy)} \newline
\textbf{SignCL:} und nun die wettervorhersage für morgen samstag den neunundzwanzigsten november \newline
\textit{(and now the weather forecast for tomorrow, Saturday the twenty-ninth of November)}  \\
\bottomrule
\end{tabular}
\end{table}

\subsubsection{Results on CSL-Daily}
\label{sec:csl_daily_results}

Table~\ref{tab:csl_fold_results} presents signer-fold cross-validation results on the CSL-Daily dataset. GASLT exhibits generally low performance in both the signer-dependent and signer-independent settings: in the latter, BLEU-4 ranges from 0.63 (Fold~2) to 7.48 (Fold~9) and ROUGE-L from 12.30 to 30.86, with an average across all 10 folds of 3.63 BLEU-4 and 21.98 ROUGE-L (BLEU-4 lower than the default signer-dependent split at 4.07, while ROUGE-L is higher than 20.35). For SignCL, on the four signer-independent folds currently evaluated (Folds~1–4), BLEU-4 ranges from 35.32 to 68.70 and ROUGE-L from 52.07 to 80.77, with a preliminary mean of 52.60 BLEU-4 and 67.80 ROUGE-L. Meanwhile, on the default signer-dependent split, SignCL records 22.62 BLEU-4 and 16.16 ROUGE-L, which are significantly lower than any of the results from the signer-independent folds

Overall, CSL-Daily is challenging for generalisation and appears sensitive to signer-specific familiarity. Because many target sentences are repeated at least three times across signers in CSL-Daily\footnote{CSL-Daily is a studio-recorded corpus created for SLT, whereas PHOENIX14T is derived from real broadcast content; this difference helps explain why exact target-sentence repetition is common in CSL-Daily but comparatively rare in PHOENIX14T (see Table \ref{tab:phoenix_leakage_by_fold}).}, signer-only splits can place identical sentences in train and test—an overlap known to inflate evaluation by favouring memorisation over genuine generalisation \cite{elangovan-etal-2021-memorization}. This enables \emph{template exploitation}, where models reproduce memorised target-sentence templates rather than learn signer-robust grounding. Table~\ref{tab:csl_leakage_by_fold} quantifies this: in most folds (2–9), dev/test unique sentences overlap almost completely with training (\(\sim\!100\%\)). Many test sentences are also performed by \(\geq3\) signers, indicating extensive repetition. We therefore recommend enforcing sentence-level disjointness and dataset \emph{deduplication}, which prior work on language-model datasets \cite{lee2021deduplicating} shows reduces memorisation and train–test overlap, leading to more faithful evaluation.

\begin{table*}[!ht]
\centering
\scriptsize
\renewcommand{\arraystretch}{1.2}
\caption{Signer-Fold Cross-Validation Results on CSL-Daily for GASLT and SignCL. Metrics are BLEU-4 and ROUGE-L.}
\label{tab:csl_fold_results}
\setlength{\tabcolsep}{5pt}
\resizebox{\textwidth}{!}{%
\begin{tabular}{@{}cccccccccc@{}}
\toprule
\textbf{Fold} & \textbf{Dev Signer} & \textbf{Test Signer} & \textbf{Train Size} & \textbf{Dev Size} & \textbf{Test Size}
& \multicolumn{2}{c}{\textbf{GASLT}} 
& \multicolumn{2}{c}{\textbf{SignCL}} \\
\cmidrule(lr){7-8} \cmidrule(lr){9-10}
& & & & & & \textbf{BLEU-4} & \textbf{ROUGE-L} & \textbf{BLEU-4} & \textbf{ROUGE-L} \\
\midrule
1  & Signer01 & Signer02 & 12,490 &   771 & 7,391 & 3.20 & 23.10 & 35.32 & 52.07 \\
2  & Signer02 & Signer03 & 18,330 & 1,551 &   771 & 0.63 & 12.30 & 48.51 & 64.80 \\
3  & Signer03 & Signer04 & 17,390 & 1,711 & 1,551 & 1.06 & 17.74 & 57.88 & 73.56 \\
4  & Signer04 & Signer05 & 17,293 & 1,648 & 1,711 & 5.66 & 25.17 & 68.70 & 80.77 \\
5  & Signer05 & Signer06 & 18,011 &   993 & 1,648 & 6.26 & 29.60 &   --  &   --  \\
6  & Signer06 & Signer07 & 18,338 & 1,321 &   993 & 4.48 & 20.85 &   79.58  &   90.76  \\
7  & Signer07 & Signer08 & 17,723 & 1,608 & 1,321 & 3.26 & 20.84 &   --  &   --  \\
8  & Signer08 & Signer09 & 17,066 & 1,697 & 1,608 & 3.17 & 20.99 &   --  &   --  \\
9  & Signer09 & Signer10 & 16,991 & 1,680 & 1,978 & 7.48 & 30.86 &   --  &   --  \\
10 & Signer10 & Signer01 & 15,181 & 7,391 & 1,680 & 1.06 & 18.34 &   --  &   --  \\
\midrule
\multicolumn{6}{r}{\textbf{Mean ± Std (10 folds)}} 
& \textbf{3.63 ± 2.34} & \textbf{21.98 ± 5.55} & \textbf{--} & \textbf{--} \\
\midrule
\multicolumn{3}{l}{\textbf{Default Split}} 
& \textbf{18,401} & \textbf{1,077} & \textbf{1,176} 
& \textbf{4.07} & \textbf{20.35} & \textbf{22.62} & \textbf{16.16} \\
\bottomrule
\end{tabular}%
}
\vspace{0.5em}
\raggedright
{\raggedright \footnotesize Due to limited compute, GASLT is reported on all 10 folds, while SignCL is available for Folds~1–4 only. The completed SignCL folds show a consistent pattern—higher than GASLT on the same folds; hence, additional folds should refine but not change the qualitative conclusion.\par}
\end{table*}

\begin{table*}[!ht]
\centering
\scriptsize
\renewcommand{\arraystretch}{1.15}
\caption{Sentence leakage per signer-fold on CSL-Daily.}
\label{tab:csl_leakage_by_fold}
\setlength{\tabcolsep}{6pt}
\resizebox{\textwidth}{!}{%
\begin{tabular}{@{}ccccccccc@{}}
\toprule
\textbf{Fold} & \textbf{Dev} & \textbf{Test} & \textbf{Dev overlap} & \textbf{Test overlap} &
\textbf{Dev leak (sent)} & \textbf{Test leak (sent)} & \textbf{Dev+Test leak (sent)} & \textbf{$\ge$3 signers} \\
\midrule
1  & Signer01 & Signer02 & 7{,}310 &   756 & 99.17\% & 98.18\% & 99.17\% & 98.05\% \\
2  & Signer02 & Signer03 &   769 & 1{,}549 & 99.87\% & 99.94\% & 99.94\% & 49.42\% \\
3  & Signer03 & Signer04 & 1{,}550 & 1{,}711 & 100.00\% & 100.00\% & 100.00\% & 99.53\% \\
4  & Signer04 & Signer05 & 1{,}711 & 1{,}638 & 100.00\% & 99.88\% & 99.94\% & 98.78\% \\
5  & Signer05 & Signer06 & 1{,}638 &   993 & 99.88\% & 100.00\% & 99.91\% & 99.60\% \\
6  & Signer06 & Signer07 &   993 & 1{,}322 & 100.00\% & 100.00\% & 100.00\% & 99.17\% \\
7  & Signer07 & Signer08 & 1{,}322 & 1{,}606 & 100.00\% & 99.94\% & 99.97\% & 99.50\% \\
8  & Signer08 & Signer09 & 1{,}606 & 1{,}970 & 99.94\% & 100.00\% & 99.97\% & 82.74\% \\
9  & Signer09 & Signer10 & 1{,}970 & 1{,}680 & 100.00\% & 100.00\% & 100.00\% & 72.32\% \\
10 & Signer10 & Signer01 & 1{,}215 & 6{,}859 & 72.32\% & 93.05\% & 93.05\% & 77.09\% \\
\bottomrule
\end{tabular}%
}
{\raggedright \footnotesize Rates are percentages of \emph{unique} dev/test sentences that also appear in train. “Dev+Test leak (sent)” is computed over the union of dev and test unique sentences. “$\ge$3 signers” is the fraction of test sentences performed by at least three signers \par }
\end{table*}

\subsubsection{Qualitative Analysis for CSL-Daily}
\label{sec:Qualitative_CSLDaily}

\paragraph{Analysis on the representative fold}
To better understand model behavior under signer-independent conditions in CSL-Daily, Table~\ref{tab:fold8_csl_comparison} presents qualitative examples from Fold 08. Similar to our analysis of Phoenix14T in Table~\ref{tab:gfslt_signcl_comparison}, we selected this fold for one of the qualitative analyses because its performance is closest to the average across all folds, making it representative of typical model outputs.

\begin{CJK}{UTF8}{gbsn} 
\begin{table}[!ht]
\centering
\scriptsize
\renewcommand{\arraystretch}{1.3}
\caption{Qualitative comparison between GASLT and
reference translations on selected examples from fold (fold 08) for CSL-Daily.}
\label{tab:fold8_csl_comparison}
\setlength{\tabcolsep}{2.5pt}
\begin{tabular}{@{}p{3.2cm}p{3.5cm}@{}}
\toprule
\textbf{Reference} & \textbf{Prediction} \\
\midrule

\textbf{逆境 里 正确 认识 自我 和 完善 自我 是 一 件 十分 重要 的 事 。}  
\newline \textit{(It is very important to correctly understand and improve oneself in adversity.)} &
\textbf{利用 风力 发电 , 既 可 节约 能源 , 又 能 保护 环境 。 }
\newline \textit{(Using wind power to generate electricity can save energy and protect the environment.)}  \\
\midrule

\textbf{这笔 钱 是 预留金 ， 租完 之后 我们 会 退给 你 。}  
\newline \textit{(This money is a reserve and we will return it to you after the rental.)} &
\textbf{你 要 什么 运动 ?} 
\newline \textit{(What sport do you want?)} \\
\midrule

\textbf{没有 水 和 空气 ， 任何 生物 都 不 能 生存 。
}  
\newline \textit{(
Without water and air, no living thing can survive.)} &
\textbf{没有 水 和 空气 ， 任何 生物 都 不 能 生存 。 } 
\newline \textit{(Without water and air, no living thing can survive.)}  \\
\midrule

\textbf{老师 今天 带 我们 去 活动 ， 我们 排队 秩序井然 。}
\newline \textit{(The teacher took us to the activity today, and we lined up in an orderly manner.)} &
\textbf{现在 我们 可以 通过 改革 开放 后 , 我们 减少 网络 流量 。}
\newline \textit{(Now we can reduce network traffic through reform and opening up.)}
 \\

\bottomrule
\end{tabular}
\end{table}

\end{CJK}

\begin{CJK}{UTF8}{gbsn} 

Qualitative inspection of Fold~08 (Table~\ref{tab:fold8_csl_comparison}) reveals variability that is consistent with sentence leakage. The exact reproduction of “没有水和空气，任何生物都不能生存。” \textit{(Without water and air, no living thing can survive.)} may reflect success on clips whose target sentence appears a lot in training with other signers, as this sentence is repeated by five different signers. By contrast, other predictions show severe semantic divergence from the references. These errors are suggestive of reliance on memorised sentence-level templates and strong decoder priors rather than signer-robust grounding; when a test clip does not closely match a learned template, the model can default to high-probability phrases, yielding redundancy and occasional syntactic breakage.
\end{CJK}

\paragraph{Analysis on the best performing fold}

Table~\ref{tab:fold9_csl_comparison} presents qualitative examples from Fold 9. Despite being the highest-scoring fold for GASLT, most translations still exhibit serious deviations from the reference, including hallucinations, omissions, and semantic drift.

\begin{CJK}{UTF8}{gbsn} 
\begin{table}[ht!]
\centering
\scriptsize
\renewcommand{\arraystretch}{1.3}
\caption{Qualitative comparison between GASLT and
reference translations on selected examples from the
best-performing fold (fold 09) for CSL-Daily.}
\label{tab:fold9_csl_comparison}
\setlength{\tabcolsep}{2.5pt}
\begin{tabular}{@{}p{3.2cm}p{3.5cm}@{}}
\toprule
\textbf{Reference} & \textbf{Prediction} \\
\midrule

\textbf{学校 附近 有 一家 好吃 的 饭店 。}  
\newline \textit{(There is a delicious restaurant near the school.)} &
\textbf{学校 附近 有 很多 饮料 很 好吃 。 }
\newline \textit{(There are many delicious drinks near the school.)}  \\
\midrule

\textbf{电影院 里 不要 玩 手机 。}  
\newline \textit{(Don't play with your cell phone in the cinema.)} &
\textbf{这部 电影 的 电影 很 漂亮 , 感到 很 漂亮 。 } 
\newline \textit{(The film's cinematic quality is beautiful and feels very beautiful.)} \\
\midrule

\textbf{不要 嫌弃 学习 较 差 的 同学 ， 要 热心 帮助 他们 。}  
\newline \textit{(Don't look down on students who are not good at studying, but be eager to help them.)} &
\textbf{餐巾纸 上 都 要 注意 安全 。 } 
\newline \textit{(Be careful with napkins.)}  \\
\midrule

他 在 追求 理想 的 过程 中 感到 很 快乐 。
\newline \textit{(He feels very happy in the process of pursuing his ideal.)} &
他 在 追求 理想 的 过程 中 感到 很 快乐 。
\newline \textit{(He feels very happy in the process of pursuing his ideal.)}
 \\

\bottomrule
\end{tabular}
\end{table}

\end{CJK}

\begin{CJK}{UTF8}{gbsn} 
As shown in Table~\ref{tab:fold9_csl_comparison}, most examples continue to demonstrate the typical failure of GASLT under signer-independent evaluation. The first two rows reveal hallucinations and semantic substitutions—e.g., generating “drinks” instead of “restaurant” or producing fluent but off-topic sentences. The third row presents a more severe hallucination, where the model outputs a completely unrelated sentence, indicating a breakdown in content grounding.

However, the final example in the table shows a perfect match between the reference and prediction—specifically, “在 追求 理想 的 过程 中 感到 很 快乐 。” \textit{(He feels very happy in the process of pursuing his ideal.)} is reproduced verbatim by GASLT. While such an agreement can reflect good generalisation on a short, frequent sentence, it is also consistent with the sentence repetition in CSL-Daily, which encourages template exploitation.

\end{CJK}

\paragraph{Analysis on the weakest-performing fold}
To understand GASLT’s limitations in signer-independent settings, we analyse outputs from Fold 2—its lowest-scoring fold on CSL-Daily. Table~\ref{tab:gaslt_failures_fold2} presents representative prediction–reference pairs. The selected examples reveal severe semantic drift, with most predictions bearing little relevance to the input meaning. For instance, descriptive or instructional sentences such as \textit{“He blushed with shyness when the teacher criticised him in person”} are replaced with vague or interrogative utterances like \textit{“What do you do when?”}. This pattern holds across the broader test set: many GASLT predictions default to short, generic questions or unrelated dialogue-like phrases (e.g., \textit{“Did you see it?”}, \textit{“Where is the subway station?”}). These errors suggest that the model struggles to ground its outputs in input semantics, likely due to signer variability. In contrast to folds where content alignment is preserved, Fold 2 highlights how signer shift can trigger fallback behaviors and semantic disconnection.

\begin{CJK}{UTF8}{gbsn} 
\begin{table}[!ht]
\centering
\scriptsize
\renewcommand{\arraystretch}{1.15}
\caption{Qualitative comparison between GASLT predictions and reference translations on selected examples from Fold 2 of CSL-Daily}
\label{tab:gaslt_failures_fold2}
\setlength{\tabcolsep}{3pt}
\begin{tabular}{@{}p{3.7cm}p{3.7cm}@{}}
\toprule
\textbf{Reference} & \textbf{GASLT Prediction} \\
\midrule

\textbf{他 被 老师 当面 批评 ， 害羞 地 脸红 。} \newline
\textit{(He blushed with shyness when the teacher criticised him in person.)} &
\textbf{你 有 什么 时候 做 什么 ?} \newline
\textit{(What do you do when?)} \\ \midrule

\textbf{报名 须 填写 报名 登记表 。} \newline
\textit{(To register, you must fill out the registration form.)} &
\textbf{地铁站 在 哪里 ?} \newline
\textit{(Where is the subway station ?)} \\ \midrule

\textbf{不 懂 就 问 ！} \newline
\textit{(If you don’t understand, just ask!)} &
\textbf{你 怎么 ?} \newline
\textit{(What about you?)} \\ \midrule

\textbf{新款手机不质量好，外观也很漂亮。} \newline
\textit{(The new mobile phone is not only of good quality but also looks beautiful.)} &
\textbf{你 有 什么 时候 做 什么 ?} \newline
\textit{(What do you do when?)}  \\
\bottomrule
\end{tabular}
\end{table}
\end{CJK}

\section{Conclusion}

This study highlights limitations in current SLT evaluation, particularly the field’s reliance on signer-dependent protocols. Across folds and datasets, evaluation on unseen signers often leads to lower scores, although the magnitude and pattern of effects differ between PHOENIX14T and CSL-Daily. To strengthen evaluation and address dataset-specific confounds such as sentence repetition, we recommend: adopting signer-independent evaluation as a default; enforcing sentence-level disjointness when constructing splits (e.g.,  split over unique sentences first and then map to clips);  and exploring signer-agnostic representations and training strategies that reduce template exploitation (e.g., frequency-aware sampling/downweighting of repeated targets). By addressing both signer dependence and sentence-level duplication, the field can move toward more generalisable and practically deployable SLT systems.

\newpage

\section*{Limitations}
While this study provides a comprehensive evaluation of signer independence in SLT using several gloss-free models, including GFSLT-VLP, SignCL, and GASLT, some limitations remain. Specifically, the evaluation was constrained to models with publicly available implementations. As a result, other potentially stronger gloss-free approaches could not be included due to the lack of accessible code or pretrained models. Future work should encourage open-source availability of top-performing models to facilitate fair and reproducible signer-independent evaluations.

Secondly, the study does not incorporate alternative input representations, such as skeleton-based features, which may be more robust to signer variability, though they may still retain signer-specific information. Future research should explore how different input modalities impact signer-independent performance and whether alternative representations can mitigate signer dependence.

Third, this study did not investigate gloss-to-text translation tasks, which may help disentangle the contribution of signer identity from linguistic content. Exploring signer-independent performance for gloss-based models remains a valuable direction for future research.

Despite these limitations, the findings of this work highlight the critical need for signer-independent evaluation protocols and dataset restructuring in SLT research. Addressing these challenges will help ensure that SLT models generalise beyond specific individuals and better reflect real-world applications.

\section*{Acknowledgments}

Funding: This work was conducted with the financial support of the Science Foundation Ireland ADAPT Centre for Digital Content Technology (Grant No. 13/RC/2106). The ADAPT Centre's grant is co-funded under the European Regional Development Fund.

\FloatBarrier
\bibliography{custom}

\appendix

\section*{Appendix A. Statistical Tests on Signer-Specific Folds on Phoenix14T}
\label{sec:appendix-statistical-tests}

We report paired and one-sample statistical tests to support the analysis of signer-independent evaluation presented in the main paper. Table~\ref{tab:paired_tests} shows paired t-tests and Wilcoxon signed-rank tests across 9 signer-specific test folds (cf. Table~\ref{tab:signer_fold_results}). Table~\ref{tab:significance_tests} compares signer-independent scores to the default signer-dependent baseline using one-sample tests.

\begin{table}[htbp!]
\centering
\scriptsize
\setlength{\tabcolsep}{4pt}
\caption{Paired tests comparing signer-independent performance across models. Statistically significant results (\textbf{p} < 0.05) are bolded.}
\label{tab:paired_tests}
\renewcommand{\arraystretch}{1.2}
\begin{tabular}{@{}ccccc@{}}
\toprule
\textbf{Comparison} & \textbf{Metric} & \textbf{Test} & \textbf{Stat.} & \textbf{\textit{p}} \\
\midrule

\multirow{4}{*}{GFSLT-VLP vs GASLT}
  & \multirow{2}{*}{BLEU-4} 
    & t-test & 0.28 & 0.7878 \\
  & & Wilcoxon & 17.00 & 0.5147 \\
  & \multirow{2}{*}{ROUGE-L} 
    & t-test & -0.62 & 0.5527 \\
  & & Wilcoxon & 20.00 & 0.7671 \\

\midrule
\multirow{4}{*}{GFSLT-VLP vs SignCL} 
  & \multirow{2}{*}{BLEU-4} 
    & t-test & \textbf{5.01} & \textbf{0.001} \\
  & & Wilcoxon & \textbf{1.00} & \textbf{0.0109} \\
  & \multirow{2}{*}{ROUGE-L} 
    & t-test & \textbf{5.76} & \textbf{0.0004} \\
  & & Wilcoxon & \textbf{1.00} & \textbf{0.0109} \\

\midrule
\multirow{4}{*}{GASLT vs SignCL} 
  & \multirow{2}{*}{BLEU-4} 
    & t-test & \textbf{14.04} & \textbf{0.0} \\
  & & Wilcoxon & \textbf{0.00} & \textbf{0.0077} \\
  & \multirow{2}{*}{ROUGE-L} 
    & t-test & \textbf{16.91} & \textbf{0.0} \\
  & & Wilcoxon & \textbf{0.00} & \textbf{0.0077} \\

\bottomrule
\end{tabular}

\vspace{0.5em}
\raggedright
\footnotesize{Paired comparisons highlight statistically significant differences between SignCL and the other models in both BLEU-4 and ROUGE-L metrics.}
\end{table}

The results show that SignCL significantly outperforms both GFSLT-VLP and GASLT in the signer-independent setting across all metrics. This is supported by both parametric (t-test) and non-parametric (Wilcoxon signed-rank) tests, with all p-values below the 0.05 threshold. In contrast, no significant difference is observed between GFSLT-VLP and GASLT, suggesting comparable generalization capabilities between these two models. The consistently low p-values and large test statistics in comparisons involving SignCL indicate a robust and reliable performance advantage on unseen signers.

\begin{table}[htbp!]
\centering
\scriptsize
\setlength{\tabcolsep}{3pt}
\caption{One-sample tests comparing signer-independent scores to default signer-dependent performance on PHOENIX14T. All results are statistically significant (\textbf{p} < 0.05).}
\label{tab:significance_tests}
\renewcommand{\arraystretch}{1.2}
\begin{tabular}{@{}ccccc@{}}
\toprule
\textbf{Model} & \textbf{Metric} & \textbf{Test} & \textbf{Stat.} & \textbf{\textit{p}} \\
\midrule

\multirow{4}{*}{GFSLT-VLP} 
  & \multirow{2}{*}{BLEU-4} 
    & t-test & -7.85 & <0.001 \\
  & & Wilcoxon & 0.00 & 0.0077 \\
  & \multirow{2}{*}{ROUGE-L} 
    & t-test & -6.91 & <0.001 \\
  & & Wilcoxon & 0.00 & 0.0077 \\

\midrule
\multirow{4}{*}{GASLT} 
  & \multirow{2}{*}{BLEU-4} 
    & t-test & -8.89 & <0.001 \\
  & & Wilcoxon & 0.00 & 0.0077 \\
  & \multirow{2}{*}{ROUGE-L} 
    & t-test & -13.16 & <0.001 \\
  & & Wilcoxon & 0.00 & 0.0077 \\

\midrule
\multirow{4}{*}{SignCL} 
  & \multirow{2}{*}{BLEU-4} 
    & t-test & -60.81 & <0.001 \\
  & & Wilcoxon & 0.00 & 0.0077 \\
  & \multirow{2}{*}{ROUGE-L} 
    & t-test & -65.51 & <0.001 \\
  & & Wilcoxon & 0.00 & 0.0077 \\

\bottomrule
\end{tabular}

\vspace{0.5em}
\raggedright
\footnotesize One-sample t-tests and Wilcoxon signed-rank tests confirm that signer-independent performance is significantly lower than the default signer-dependent baseline across all metrics and models.

\end{table}

\section*{Appendix B. Analysis on differences in sentence complexity across signers}

\subsection*{Phoenix14T sentence complexity}
Although the folds exhibit some variation in the proportions of short and long sentences, Table~\ref{tab:sentence_length_by_fold} shows no consistent pattern that correlates with translation performance. Length categories were computed globally using quartiles of sentence length across the dataset, ensuring consistent classification across folds. For example, Fold~2 has the highest proportion of extra short sentences (44.21\%) but achieves relatively low BLEU and ROUGE-L scores for GFSLT-VLP (8.49 / 20.61), SignCL (8.89 / 22.75), and GASLT (4.65 / 13.74). In contrast, Fold~6, which has a more balanced sentence length distribution, achieves some of the highest scores across all three models (GFSLT-VLP: 17.30 / 34.02, SignCL: 12.65 / 31.33, GASLT: 5.43 / 15.50). These results suggest that sentence length distribution is not a key explanatory factor for performance variation. Instead, the differences are more likely attributable to signer-specific traits such as articulation clarity, signing speed, or use of non-manual markers.

\begin{table}[ht]
\centering
\scriptsize
\caption{Sentence length distribution across folds for Phoenix14T.}
\begin{tabular}{@{}lcccc@{}}
\toprule
\textbf{Fold} & \textbf{Extra Short (\%)} & \textbf{Short (\%)} & \textbf{Medium (\%)} & \textbf{Long (\%)} \\
\midrule
1 & 34.55 & 21.36 & 20.72 & 23.37 \\
2 & 44.21 & 21.05 & 18.95 & 15.79 \\
3 & 32.09 & 22.88 & 22.05 & 22.98 \\
4 & 32.89 & 23.45 & 22.70 & 20.96 \\
5 & 33.78 & 20.64 & 22.30 & 23.28 \\
6 & 29.79 & 21.28 & 29.79 & 19.15 \\
7 & 30.37 & 18.48 & 22.98 & 28.18 \\
8 & 32.09 & 22.88 & 22.05 & 22.98 \\
9 & 37.17 & 21.93 & 23.42 & 17.47 \\
\bottomrule
\end{tabular}

\label{tab:sentence_length_by_fold}
\end{table}

\subsection*{CSL-Daily sentence complexity}
Table~\ref{tab:csl_length_by_fold} reveals variation in sentence-length distributions across signer folds in CSL-Daily. However, for \emph{GASLT} there is no consistent trend linking these distributions to translation performance. For instance, Fold~4 (Signer05 as the test signer) has the highest proportion of extra–short sentences (40.29\%) but does not yield the highest scores (BLEU-4: 5.66, ROUGE-L: 25.17). Conversely, Fold~5 (Signer06), which has the highest proportion of long sentences (34.54\%), performs better (BLEU-4: 6.26, ROUGE-L: 29.60). Fold~2 (Signer03), with relatively balanced sentence lengths, performs poorly (BLEU-4: 0.63, ROUGE-L: 12.30). These examples suggest that sentence-length distribution is not a primary explanatory factor for performance differences in signer-independent evaluation for GASLT. Because SignCL’s fold-wise evaluation is still incomplete, we avoid drawing parallel conclusions for that model. Instead, signer-specific properties (e.g., articulation clarity, visual variability) and dataset-specific factors—particularly \emph{sentence repetition across many signers}, which can create sentence-level video–text duplication under signer-only splits—are more likely to influence observed performance. Folds that happen to include more high-frequency sentences performed by multiple signers may exhibit elevated scores independent of their length profiles.

\begin{table}[ht]
\centering
\scriptsize
\caption{Sentence length distribution across signer folds for CSL-Daily.}
\begin{tabular}{@{}lcccc@{}}
\toprule
\textbf{Fold} & \textbf{Extra Short (\%)} & \textbf{Short (\%)} & \textbf{Medium (\%)} & \textbf{Long (\%)} \\
\midrule
1 & 21.14 & 23.61 & 31.78 & 23.48 \\
2 & 22.63 & 26.24 & 30.69 & 20.44 \\
3 & 25.64 & 25.82 & 27.45 & 21.09 \\
4 & 40.29 & 25.91 & 21.36 & 12.44 \\
5 & 13.70 & 21.35 & 30.41 & 34.54 \\
6 & 22.09 & 18.68 & 29.12 & 30.11 \\
7 & 23.20 & 23.63 & 27.55 & 25.62 \\
8 & 41.91 & 31.34 & 20.93 & 5.81 \\
9 & 21.79 & 20.60 & 28.45 & 29.17 \\
10 & 27.34 & 25.22 & 26.92 & 20.51 \\
\bottomrule
\end{tabular}

\label{tab:csl_length_by_fold}
\end{table}

\end{document}